# A Robotic Cyber-Physical System for Automated Reality Capture and Visualization in Construction Progress Monitoring

Srijeet Halder[a,b], Kereshmeh Afsari[b], Abiola Akanmu[b]

[a] Department of Sustainable Technology and the Built Environment, Appalachian State University, Boone, NC, United States. Email: halders1@appstate.edu. Corresponding author.
[b] Myers Lawson School of Construction, Virginia Tech, Blacksburg, VA, United States.

**Abstract**

Effective progress monitoring is crucial for the successful delivery of the construction project within the stipulated time and budget. Construction projects are often monitored irregularly through time-consuming physical site visits by multiple project stakeholders. Remote monitoring using robotic cyber-physical systems (CPS) can make the process more efficient and safer. This article presents a conceptual framework for robotic CPS for automated reality capture and visualization for remote progress monitoring in construction. The CPS integrates quadruped robot, Building Information Modelling (BIM), and 360° reality capturing to autonomously capture, and visualize up-to-date site information. Additionally, the study explores the factors affecting acceptance of the proposed robotic CPS through semi-structured interviews with seventeen progress monitoring experts. The findings will guide construction management teams in adopting CPS in construction and drive further research in the human-centered development of CPS for construction.

**Keywords**: robotics; building information modelling; progress monitoring; quadruped robot; human-robot interaction;

## 1. Introduction

Construction progress monitoring is essential in ensuring timely project completion [1,2]. Progress monitoring involves the collection of project progress data (e.g., site images) through inspections, analysis of the collected data to identify discrepancies, and communication of the results in a form that can be easily interpreted by various project stakeholders [1]. However, construction progress monitoring is a complex and challenging process that requires the integration of various data sources and stakeholders [3]. One of the main challenges of construction progress monitoring is the lack of accurate and timely data, which is important for effective progress monitoring [3,4]. Another challenge is the fragmented nature of the construction industry, regular inspections of a construction project are undertaken by various project stakeholders such as owners, project managers, architects, engineers, contractors, subcontractors, end users, and facility managers, who use different methodologies and documentation system for monitoring the project [5]. This can lead to inconsistencies in data collection and analysis, making it difficult to gain a complete picture of project progress [1]. Finally, manually inspecting the job site can be time-consuming and prone to errors, which can impact the accuracy and frequency of progress monitoring [1,2]. From a safety standpoint, project stakeholders inspecting construction sites are also exposed to various safety hazards, e.g., falls [6]. To address the above challenges, the construction industry must embrace digital technologies that can integrate data from various sources and provide real-time insights into project progress [7].

Cyber-physical systems (CPS) have emerged as a potential solution to address the challenges of construction progress monitoring [7,8]. A robotic CPS integrates mobile robots with sensing, and computing technologies to create an intelligent, autonomous, and mobile system for a variety of applications [9]. Robotic CPS equipped with sensors, cameras, and other technologies that allow for real-time data collection and analysis can improve the efficiency and completeness of progress monitoring [7]. By implementing robotic CPS, construction companies can enhance safety, reduce delays, and improve overall project outcomes. While past research has explored using robots for autonomous inspection of construction sites, less research has been done on understanding the industry acceptance of the technology.

The main contribution of this research is the development of a conceptual framework for robotic CPS for automated reality capture and visualization in construction progress monitoring through the integration of quadruped robots, Building Information Modelling, and 360° reality capture. Additionally, the acceptance of the proposed CPS framework by the industry is studied through semi-structured interviews with seventeen industry experts involved in the construction progress monitoring work. The semi-structured interviews are used to identify the factors affecting the acceptance of the proposed CPS for remote progress monitoring. The article also discusses the implications of the proposed system for the future of construction technology and its potential impact on the industry. The following research questions are answered in this research:

- **RQ1** – What conceptual framework for robotic CPS can be applied to address the inefficiencies of conventional inspection in construction progress monitoring?
- **RQ2** – To what extent is the proposed conceptual framework for robotic CPS acceptable as an alternative means to traditional construction progress monitoring?

The scope of the current work includes developing a conceptual framework for robotic CPS for automated reality capture and progress monitoring, then implementing a prototype to validate the framework through experimental investigations, and finally validating the proposed conceptual framework with construction experts. Advancement of the robot navigation and locomotion system. This study also makes an assumption that adequately accurate and complete BIM is available for construction projects for application of the proposed CPS. Improving accuracy of BIM in practice and increasing adoption of BIM is not under the scope of current work. This work also builds upon previous studies by the authors [10–14].

The structure of the rest of the paper is as follows. In the Background section, previous work on the topic is discussed. Next, the Methodology section describes the research methodology used in this study to develop and evaluate the framework for the proposed robotic CPS for construction progress monitoring. Next, the implementation details of the proposed robotic CPS are explained. The Results section describes the data collected for evaluating the robotic CPS through an experimental investigation and expert validation interviews. The Discussion section explains the study's implications for the future of construction progress monitoring with the proposed robotic CPS. Finally, the study is concluded in the Conclusion section with a discussion on the limitations of the current work and recommendations for future research.

## 2. Background

### 2.1. Automated Inspection and Monitoring in Construction

The process of construction progress monitoring includes a set of inspection tasks that involve many subtasks, including conducting walkthroughs of the job site, collecting visual and other information regarding the as-built status, processing the collected data, and extracting and communicating the information regarding the inspection results with project stakeholders [1]. This process is repeated for the duration of construction. Some inspection tasks are difficult to perform for humans due to inaccessibility, e.g., in confined spaces like inside air-conditioning ducts [15,16], water-filled pipelines and tunnels [17], offshore structures [18], and small spaces in the wall [19]. Some situations can be hazardous for humans too, such as inspections at height [20], structures that are subject to natural disasters [21], or a bridge deck [22]. Importantly, real-time monitoring of construction projects is the key to keeping pace with the construction progress and reducing rework [23]. However, frequent monitoring of construction projects also requires significant time, manpower, data management, and travel to multiple job sites [23]. Automating the data collection process in progress monitoring can significantly affect the management of the project [24], can prevent schedule delays and cost overruns, and can improve the overall quality of construction work [25]. Data collected through traditional inspection methods is also unordered which requires additional effort to review and analyze the progress and describe the discrepancies to other project stakeholders [3].

Due to these challenges associated with manual construction progress monitoring, many researchers have focused on introducing automation in the progress monitoring process [26,27]. For example, drones, and ground robots have been used to collect images and videos from construction job sites [28,29]. Semantic modeling and natural language processing (NLP) has been used to analyze construction documents for code compliance checking [30]. Similarly, photogrammetry has been used by Golparvar-Fard et al. [24] using unstructured site photographs for processing site progress data. The information extraction step may include taking measurements [31], calculating activity percentage completion [32], and finding discrepancies [33] from images or point clouds. Studies have also developed computer vision models to detect cracks in concrete and missing electrical outlets [34,35]. Convolutional Neural Network (CNN) is a widely-used computer vision technique for defect detection and object identification [35–37]. For communication purposes, web-based collaboration tools, such as Procore, BIM360, CoConstruct, PlanGrid, eBuilder, and Aconex are used to record and communicate post-inspection details [38]. These techniques remove the subjectivity of human inspectors and reduce the time for construction inspection and monitoring [24].

BIM is an important tool that has the potential to support automation in many aspects of construction [39,40]. Through BIM, a building model is constructed digitally and loaded with various data that can support decision-making and analysis throughout the building lifecycle [39]. A BIM model stores both geometric (shapes, dimensions, etc.) and semantic (object type, material, relationships, schedule, etc.) information about the building elements in a structured format [40]. This geometric and semantic information can be used, for example, to guide robots on a construction site [41] as well as compare as-built information with the as-planned information [42] in the process of construction progress monitoring.

Another important automation tool for construction is robots that have been used as primary data collection agents in the robotic inspection of construction [43–45]. Robotic construction inspection involves the use of automated or semi-automated robots, for conducting a partial or full inspection of construction projects. Mobile robots, which are capable of moving from one place to another, can be categorized into different classes based on their locomotion, such as legged robots [10,11,46], wheeled robots [47,48], unmanned marine vehicles [49,50], unmanned aerial vehicles [51,52], etc. A robot's locomotion is the method of movement that the robot uses to move from one place to another [53]. Studies have used many types of commercial robots like Husky with Kinova Arm [54] and Clearpath Jackal [43], and Boston Dynamics Spot [11,55]. They have also developed custom-built robots for specific purposes, such as magnetic legs [56], rope climbing robots with rotors [57], wheeled robots with propellors [44], and wall-sticking drones [58].

In this research, robots are used as autonomous data collection agents to facilitate construction progress monitoring. Robots are considered one of the components of the CPS. A CPS utilizing a robot as a core component is referred to as robotic CPS [59–61]. As per the taxonomy used by [62], the robot in this research is used in the sensing and device layer of the CPS.

## 2.2. Robotic Cyber-Physical Systems in Construction

Cyber-Physical Systems (CPS) are integrated networks of computing and physical components with the ability to sense the real environment and interact with it through sensors and actuators [8,63]. The term CPS was coined in 2006 and has been identified as an influential research area by the US National Science Foundation (NSF) in 2008 [64]. A CPS can introduce autonomous capabilities to physical processes by autonomously adapting to changing environments and user needs [65]. CPS has found application in many industries including automotive, manufacturing, power generation, and construction [65,66]. Humans play an important role in providing feedback to the CPS to handle complex tasks in an unstructured environment by combining human cognitive skills with the autonomous behavior of CPS [65]. CPS using humans as part of the feedback loop is called human-in-the-loop CPS, abbreviated as HiTL-CPS or HiLCPS [65,67–69]. According to the taxonomy proposed by [69], humans play one of the three roles in HiLCPS – data acquisition, state inference, and actuation. In other words, humans may gather input data, support decision-making, or interact with the physical environment. The physical-side components sense the environment and generate data for processing by the cyber-component, which in turn provides feedback and control to the physical components [70]. Figure 1 shows the conceptual model of CPS and related concepts.

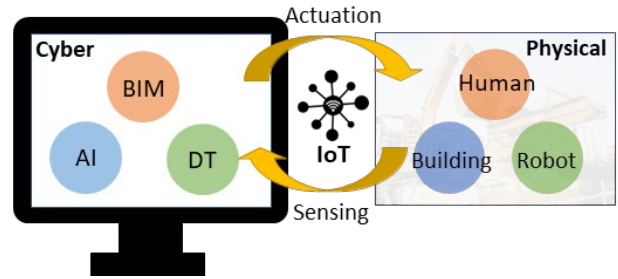

Figure 1 Conceptual model of CPS and related concepts (DT-Digital Twin; AI-Artificial Intelligence; BIM-Building Information Modelling)

In construction, Yuan and Anumba [71] used CPS for temporary structure monitoring to prevent accidents from structural failures in scaffolding and other temporary structures. Fang et al. [72] developed a CPS framework for intelligent crane operations to assist the crane operator in the safe operation of cranes and avoid crane-related accidents. Eskandar et al. [68] proposed a HiLCPS framework for construction safety monitoring. They identified four different roles of humans in a HiLCPS – data gathering, state inference, actuation, and feedback. Akanmu et al. [62] identified four potential applications of CPS in construction – progress monitoring and control, change management, as-built information, and operation and maintenance. Akanmu et al. [62] also identified six different requirements for CPS to be useful in construction – real-time, adaptability, networked, predictability, human-in-the-loop, and reliability. According to [62], CPS is composed of four layers of abstraction – sensing and device layer, communication layer, application layer, and actuation layer as shown in Figure 3 Layers of CPS [59]. The bi-directional interactions between the components of the CPS are mediated by sensors and actuators [8]. The sensors perceive the physical systems and transcode and transfer the information to the cyber components, while the actuators transfer the required actions from the cyber components to the physical components.

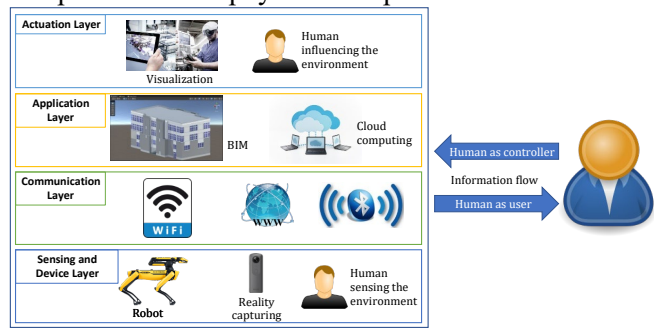

Figure 2 Layers of CPS [59]

Many supporting technologies for CPS have been identified by Linares et al. [73], which may serve in one of the four layers of CPS identified by [62]. These technologies can be classified as the physical or cyber-side components of the CPS. Some examples of such supporting technologies are presented in Table 1. From the available literature on these components, some of the key components that can be applied within a CPS framework are robots, BIM, reality capture devices, and cloud. These components and their potential in construction progress monitoring are discussed in the following subsections.

Table 1 Supporting technologies for CPS [73]

| Technology | Classification |
| --- | --- |
| Computer-Aided Design (CAD) | Cyber |
| Building Information Modeling (BIM) | Cyber |
| Augmented/Virtual/Mixed Reality (AR/VR/MR) | Cyber |
| Artificial Intelligence (AI) | Cyber |
| Blockchain | Cyber |
| Global Position System (GPS) | Physical |
| Portable devices (e.g., smartphones, tablet pc) | Physical |
| Internet of Things (IoT) devices | Physical |
| Drones/Robots | Physical |
| 3D Printers | Physical |
| Exoskeletons | Physical |
| Autonomous Vehicles | Physical |

### 2.3. Robots

Data collection task in the construction progress monitoring process is time-consuming and involves large overhead costs [1,74]. The overheads are related to productive hours spent walking through the site by multiple stakeholders, costs associated with traveling between the sites, and providing safe access to the stakeholders to the inspection areas [1,74].

Robots provide a more efficient alternative to manual data collection tasks in construction progress monitoring [26,27]. Robots are mechanical devices with a certain degree of automation [75]. The control of robots may vary from teleoperated (completely manual) to fully automated (least human intervention) [76]. Fully automated robots can be programmed to collect specific data on-demand from target locations with the least human involvement [43–45], which can save a significant amount of productive hours as well as travel costs.

In addition to the economic point of view, robots also provide safe access to places that might be hazardous or inaccessible to human inspectors, such as at-height inspections [20], structures subjected to natural disasters [21], or a bridge deck [22]. Various forms of robots have been developed by past researchers for inspections in these locations, e.g., legged robots [10,11,46], wheeled robots [47,48], unmanned marine vehicles [49,50], unmanned aerial vehicles [51,52], etc. Therefore, robots are considered an important supporting technology for CPS in construction progress monitoring in this research.

### 2.4. BIM

BIM is an information-rich digital representation of the building that has introduced a paradigm shift in the construction industry in terms of how buildings are designed, constructed, and operated [77]. The BIM may contain geometry, materials, textures, schedule, and cost as well as many other physical and engineering properties of the building elements, e.g., thermal capacity, conductivity, etc. that can be used for many documentation and analytical purposes [78]. In addition, it may also contain semantic information about the building elements (e.g., element type) as well as logical relationships between the elements [12,79]. It's been reported that utilizing BIM can provide 25-30% productivity improvement [7]. BIM has been used with robotics in previous literature. The geometrical information of the building stored in BIM can be used to navigate the robot across the building [12]. BIM also provides an intuitive method to visualize the building [80–82].

Due to these reasons, BIM is an important supporting technology of CPS that will be used in this research as one of the CPS components. The BIM will provide essential prior knowledge about the building environment to enhance the automation capability of the robot. The BIM will also be used to create an intuitive 3D environment for the user of the CPS to set goals and objectives (also called mission) for the robot as well as to visualize the construction progress captured by the robot using reality capture technologies.

### 2.5. Reality capture

Reality capturing is a set of techniques that are used to create a digital replica of real-world scenarios [83,84]. It includes 360° camera technology [83], laser scanning [84], and photogrammetry [85]. Reality capturing provides vicarious access to real-world information in the form of 2D or 3D visual data [83]. Reality capture has been used in past literature to "bring the site" to project stakeholders [86]. Reality capturing along with immersive virtual reality technology provides an enhanced sense of presence to a remote user not physically present at the site [86]. Due to the potential of reality capturing in construction progress monitoring evidenced in past literature, it has been considered an important component of the proposed CPS in this research. However, reality capturing using only 360° cameras is explored in this study.

### 2.6. Cloud computing

Cloud computing is the use of remote computing infrastructure separated from the user [87]. Cloud computing provides a scalable network infrastructure to connect devices across the globe [13]. Cloud can be used for remote computing to avoid the requirement of a high-performing on-site computation system [88]. It can also be used as a database to store data collected from the site using sensors and/or cameras [71]. Cloud is useful to provide multi-user access to the project information for geographically separated users [87]. In this research, the cloud is utilized as a communication tool to facilitate information exchange between the project site and remote project stakeholders. It facilitates the storage of information, multi-user access, and security control (by authorizing the user before establishing a connection with the on-site robot). Google Cloud Platform (GCP) and Amazon Web Server (AWS) are two of the scalable cloud computing platforms commonly used in the literature [13,71]. Due to the author's previous experience in working with GCP [13], in this study GCP will be used for cloud computing.

## 2.7. Integration of the CPS components

Researchers have found many opportunities for CPS in the construction industry, including, improvements in efficiency, flexibility, reliability, autonomy, resilience, and remote access [89]. While each of the CPS components or supporting technologies of the CPS discussed above provides significant improvement in the progress monitoring process, their integration is expected to add more resilience [89]. While robots are excellent data collection tools, they lack the versatility and adaptability of humans. Similarly, BIM and cloud provide mechanisms to store and exchange rich information about the project, yet they are disconnected from the real world in the sense that changes in the real world are not automatically reflected in the BIM or cloud. Integration of these cyber and physical components in a CPS through a synergistic relationship between each is expected to add more robustness and efficiency that can significantly improve the progress monitoring process. Such integration will also improve coordination and collaboration [90].

## 2.8. Technology Acceptance Model (TAM)

Technology adoption and diffusion have been investigated by many researchers [91–94]. The Technology Acceptance Model (TAM) proposed by Davis [91] made a significant contribution to the theory and analysis of technology use and adoption [93]. TAM was developed to model and predict the user acceptance of an information system [92]. Davis investigated and examined the reasons behind people's acceptance of information technology [93]. In TAM, an individual's behavioral intention, which is influenced by their attitude, beliefs, and other factors, determines their actual behavior [93]. The TAM was preceded by the theory of reasoned action (TRA) and the theory of planned behavior (TPB) [93]. Based on the TRA from the field of social psychology, Davis developed the TAM to explain and predict the actions of IT users.

According to TAM, perceived usefulness and ease of use are the primary factors influencing attitude and intention to use technology. All other factors indirectly affect intention and attitude through perceived usefulness and ease of use [92,93]. Thus, as illustrated in Figure 4, the factors known as perceived usefulness and perceived ease of use are the fundamental components of TAM. The perceived usefulness of a technology measures how much users think utilizing it enhances their capacity to do their work, while the perceived ease of use measures how much they think technology is easy to use or free of exertion [92,93]. A technological system may be easy to use but not useful in terms of achieving work objectives. Conversely, the system may be useful but not usable [95].

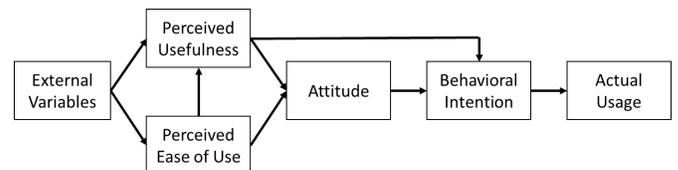

Figure 4 Technology Acceptance Model adopted from Park and Park [93]

Sorce and Issa [96] tested the original TAM for the adoption of Information and Communications Technology (ICT) in the U.S. construction industry. They found that the TAM is a valid model that reflects the technology adoption of ICT in the US construction industry, i.e., the perceived usefulness and perceived ease of use are reliable predictors of the accepted use of technology. Therefore, in the current research, the original TAM will be used as an evaluation tool in the evaluation of the proposed CPS framework.

## 2.9. Research Gaps

Even though there has been some research on CPS in construction in the past [7,62,68,71,72], robot control has not been integrated before with BIM for automated reality capture and 3D visualization. Also, previous studies on robotic inspection of construction have not studied the technology acceptance by the industry. Therefore, this research developed a framework for robotic CPS for automated reality capture and 3D visualization and evaluated the framework from the industry acceptance point of view.

## 2.10. Theoretical Underpinning

This work is anchored in the Technology Acceptance Model, as explained in section 2.8, which is used as the basis for the evaluation of the proposed framework for robotic CPS. Technology Acceptance Model has been a widely accepted model to explain the acceptance of technology in the industry. Most research on CPS in construction has ignored the acceptance of the system from the industry point of view, which lead to slow adoption of the technology in the construction industry. Therefore, this research specifically focuses on the acceptance of the proposed CPS.

## 3. Methodology

This study adopted a qualitative approach to developing and evaluating a robotic CPS for construction progress monitoring. A computational framework for integrating BIM, reality capture, and robot control systems is developed to automate the process of reality capture and visualization in a 3D virtual environment. The components of the robotic CPS are identified from the literature and the previous work of the authors and integrated into a fundamental CPS framework. The proposed framework is then evaluated by developing a proof-of-concept prototype. The prototype is then evaluated in an experimental setup. Then, the developed robotic CPS is validated through expert interviews.

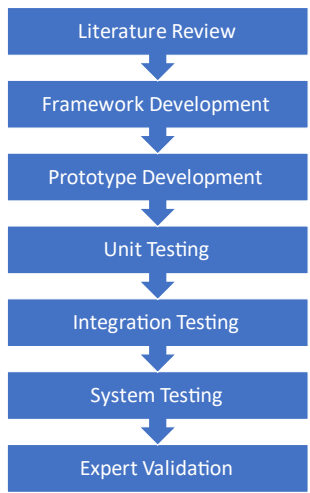

Figure 5 Overview of Research Methodology

### 3.1. Framework Development and Implementation

The robotic CPS framework was designed to enable automated reality capture, processing, and visualization of as-built information for construction progress monitoring in real-time and facilitate communication between project stakeholders located remotely. The development of the framework involved several steps. First, the technical components of the system were identified, including the robotic hardware, reality-capture devices, and software applications. Second, a software application was developed to enable the integration of the identified components and automate the data capture and visualization workflow. The software application was also designed to provide a 3D virtual environment for visualizing the captured data. Third, unit testing was performed on individual modules of the system as shown in  and described in the Proposed Framework section by providing sample input and verifying the outputs. Fourth, integration testing was performed by evaluating the communication between the individual modules by analyzing the runtime logs of each of the modules. Finally, system testing was performed by running the whole system through experiments as explained in section 3.2.

### 3.2. Experimental Investigation

The developed robotic CPS was evaluated in the developed prototype by conducting experimental investigations in the Bishop-Favrao Hall (BFH) at Virginia Tech. The tests were conducted by importing the BIM model of the building into the system and running the system to collect 360° pictures autonomously and finally, visualizing the 360° pictures in a 3D virtual environment. Six different locations were selected, where the robot captured 360° images autonomously. The experimental path used for testing is shown in Figure 6.

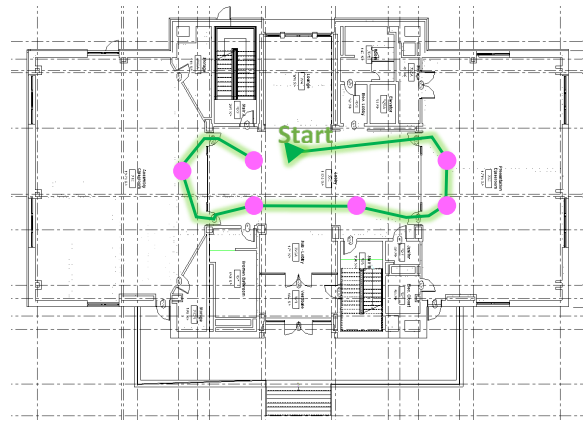

Figure 6 Experimental path for testing the framework prototype in BFH

### 3.3. Expert Validation

Expert validation was performed for the proposed robotic cyber-physical system that involved conducting semi-structured interviews with industry experts with expertise in construction progress monitoring work. A similar approach to [97] was adopted in this study, in which construction experts were interviewed one-on-one and then were asked to fill out a short questionnaire survey online as explained later in this section.

The experts were selected based on their experience and expertise in the field of construction progress monitoring through convenience and snowball sampling, i.e., experts from the professional network of the authors were first approached to participate in the study, who then introduced their colleagues. The eligibility criteria were set as work experience in the construction industry in the U.S. and a self-rated familiarity level of 3 with construction progress monitoring on a scale of 1 to 5 (1 not being familiar at all and 5 being completely familiar). The experts were shown and explained the developed robotic CPS during about an hour-long meeting and demonstration conducted online through Zoom teleconferencing software and the participants were allowed to use the system remotely. They could only use the user interface for defining missions for the robot and observing pre-recorded images. A semi-structured interview was conducted afterward to gather their feedback and opinions on the developed robotic CPS. The interview questions were designed to cover the different aspects of the system, including its technical components, usefulness, ease of use, and factors for successful adoption. The questions for the semi-structured interview are provided in Appendix A. The interview protocol was reviewed and approved by Virginia Tech IRB (#22-428) and was piloted with one test subject whose data was not included in the results and analysis.

The interview data were analyzed using the thematic content analysis method using NVIVO qualitative analysis software to identify recurring themes and patterns in the experts' responses. The analysis provided insights into the experts' acceptance of the developed CPS and its potential for adoption in the construction industry. At the end of the interviews, the experts were given a short online questionnaire survey using the Question Pro survey tool licensed by Virginia Tech. The questionnaire measured the perceived usefulness and perceived ease of use of the system by the study participants and was based on the TAM [91] and recommendations provided by Lewis [98].

## 4. Proposed Framework

This section explains the proposed framework for automated reality capture and visualization using a robotic CPS. An overview of the workflow of the robotic CPS is shown in Figure 7. First, the 3D model of the building is imported from Revit as an FBX file to create a 3D virtual environment. This is a one-time process unless the BIM model is updated. The FBX file includes information about the geometry, textures, and layers of the structure from the BIM. The geometry and layers are used to navigate the robot. While the geometry provides the dimensions, the layers provide semantic information about the element types, i.e., floors, doors, walls, etc. This information is used to decide what surfaces are walkable for robot path planning. The textures are useful in creating realistic rendering for visualization.

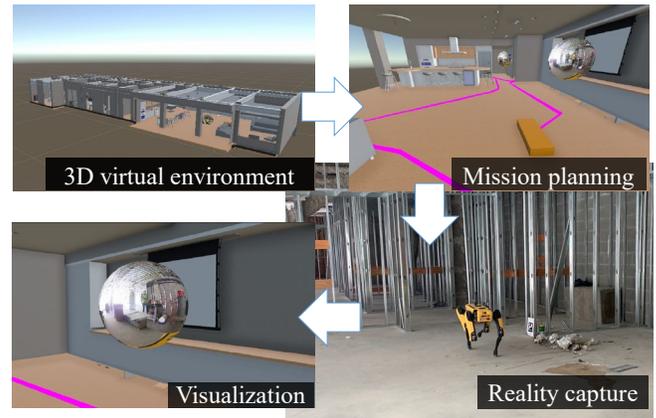

Figure 7 Workflow of the proposed robotic CPS

Anumba et al. [454] suggested a four-layered architecture for CPS as shown in Figure 8. The CPS proposed in this research is based on a similar four-layered architecture. In the sensing and device layer, a robot and a reality capturing tool collect data from the site (as-built information). The communication layer comprises of internet and Wi-Fi access points that encode, transmit, and decode data between other components of the CPS. The application layer comprises of BIM to store the geometry (dimensions, orientation, position, etc.), materials, textures, and element types of the building elements. This layer also comprises of the AI models for path finding and cloud server for data persistence. Finally, the actuation layer consists of user interface (UI) and the human for combining the as-built and as-planned information and decision-making.

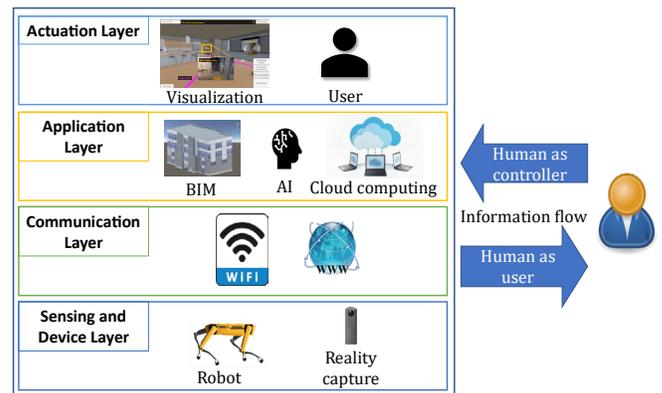

Figure 8 Four layered CPS architecture [454]

Figure 9 shows the detailed workflow in the proposed framework. The user can navigate the 3D virtual environment and select locations where reality capture should be performed by the robot. These points are referred to as Discrete Reality Points (DRP) in this article. The middleware localizes the robot in the building by querying the robot's status from the robot's Application Programming Interface (API) and sends it back to the Cloud Server. The Web Interface requests the robot's location from Cloud Server and automatically calculates the optimum path from the robot's current location to all the DRPs created by the user and packages it as a mission for the robot's autonomous navigation. Upon receiving the mission data from the Web Interface through the Cloud Server, the middleware drives the robot autonomously and captures 360° images at the specified DRPs. Finally, the captured images are downloaded from the camera and sent back to the Web Interface through the Cloud Server to be visualized within the 3D virtual environment. The user can then perform a virtual walkthrough of the site in the 3D virtual environment, which displays both as-built and as-planned conditions. The following subsections explain the whole workflow in more detail.

Based on the layered architecture of CPS suggested by [62] and depicted in , the proposed CPS framework comprises distinct layers, namely, sensing layer, communication layer, application layer, and actuation layer. The sensing layer consists of the reality capture device responsible for capturing visual information from the construction site. The communication layer utilizes WiFi and the internet to facilitate the exchange of user commands and site data between various components, including the user, cloud, middleware, robot, and camera. The application layer consists of a web interface that enables the visualization of the facility's 3D model alongside the reality captures obtained from the site as well as the cloud server to store and relay the user commands and site data between the site and user locations. Lastly, the actuation layer consists of a mobile robot equipped with a camera, which provides mobility to the system and facilitates the capturing of visual data from specific areas of interest for progress monitoring.

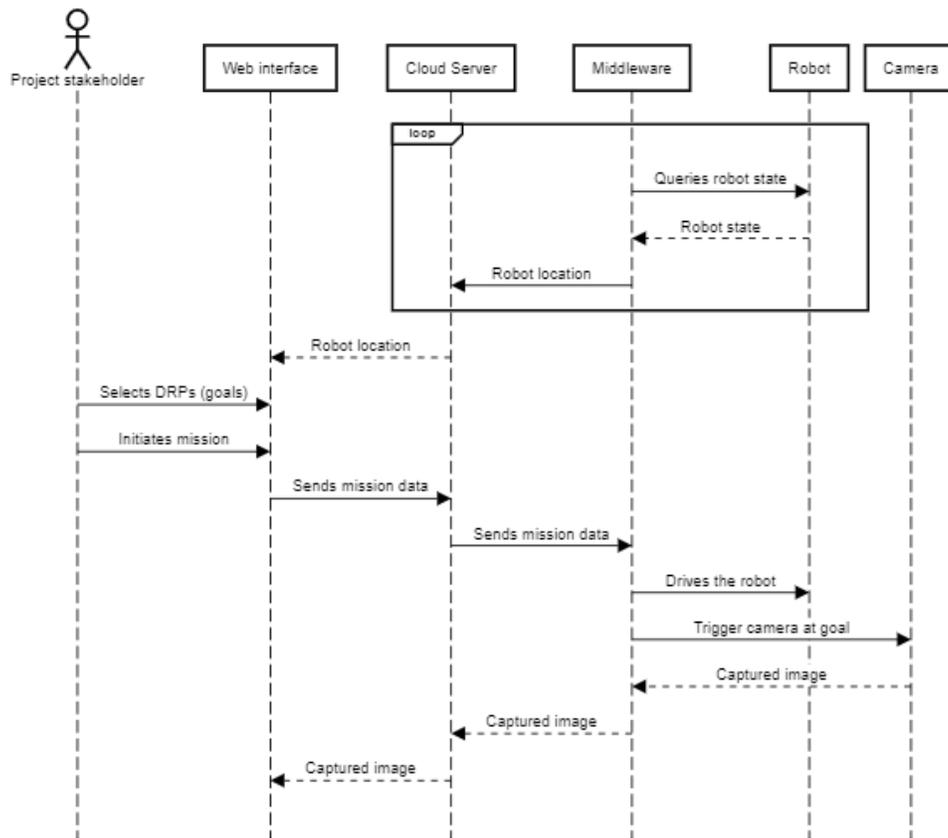

Figure 9 Proposed framework for reality capture and visualization for progress monitoring using a robotic CPS

## 5. Prototype Implementation

In this section, the implementation details of the proposed framework into a proof-of-concept prototype is explained. The prototype is developed in a modular architecture in Unity and Python as explained below. Subsection 5.1 explains the system architecture and logical flow through the system. Subsections 5.2 to 5.5 explain each of the steps of the workflow in the proposed framework shown in Figure 7.

### 5.1. Overview of system architecture

Figure 10 shows the architecture of the CPS developed in this study. There are five main physical components in the CPS:

a) Client Interface – This is a computer used by the user to log into the system.
b) Middleware – This is another computing device situated at the site which controls the robot. It can be a server located at the project office and connected to the local network at the site or an edge computing device like Raspberry Pi installed on the robot.
c) Robot – This is the quadruped robot that navigates the job site and carries the 360° camera to capture images. It must be connected to the same network as the middleware.
d) Camera – This is the main reality capture device installed on the robot and must also be connected to the same local network as the middleware and the robot.
e) Fiducials – These allow linking the BIM with reality (i.e., cyber components and physical components). These are printed on paper and installed across the site for robot localization.

The proposed CPS also comprises four cyber components as follows:

a) Web Interface – The Web Interface is the main interface between the user and the CPS. The Web Interface is an HTML page served from a Cloud Server hosted on Google Cloud Platform (GCP)
b) BIM – The BIM is a key component of the CPS. The BIM provides the geometry and textures of different elements of the building being constructed. This helps in the visualization of the project information. The BIM also provides semantic information about the object class, e.g., doors, floors, walls, and furniture. This information is used to plan the path of the robot by avoiding furniture, walls, building edges.
c) Cloud – The Cloud Server hosted on the GCP provides both computation and storage capabilities to the CPS. The Cloud Server serves the Web Interface to the user when requested and also relays the input and images between the Middleware and the Web Interface. In addition to that, the Cloud database also stores the images captured by the robot and sends them to the Web Interface when the user selects a specific date.
d) Artificial Intelligence (AI) - Artificial Intelligence refers to the ability of computer systems to make inferences from a set of information. In the proposed CPS, the A* heuristic method is used for path planning implemented through Unity's NavMesh package [99].

The workflow of the CPS starts from the system localizing the robot in the 3D virtual environment. Then the robot path is calculated and sent over to the middleware for execution. Middleware controls the robot and captures 360° pictures and sends back to the client interface for visualization. The user then navigates the 3D virtual environment to inspect the site remotely and perform progress monitoring. These steps are explained in detail in the following subsections.

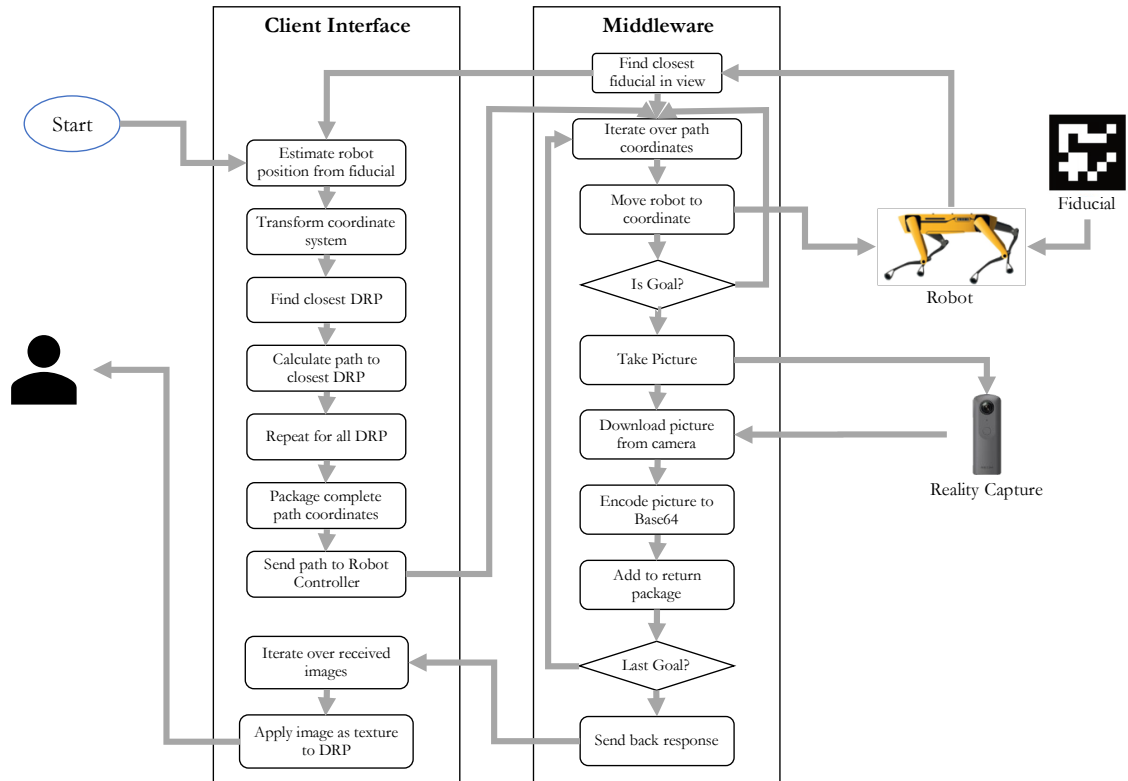

Figure 11 System architecture

## 5.2. 3D virtual environment

The 3D virtual environment is created in Unity, which is a popular game engine also used in research for VR/AR

development as well as simulations [87,100,101]. The 3D virtual environment was created by importing the FBX file generated from Revit as a prefab in Unity. In this step, it is assumed that the vertical direction of the building model aligns with the vertical direction in the Unity coordinate system. Otherwise, the model needs to be rotated. Directional lighting is separately added to the virtual environment since the FBX file does not contain lighting information. Mesh collider is applied to all the layers except the door layer. Mesh collider prevents the user from passing through the object. A digital twin of the robot is added to the virtual environment as a cuboid. A backend script receives the location of the robot from real life through the cloud server and updates the position of the robot's digital twin in the virtual environment as the robot in the real world moves. The virtual environment was designed with the metaphor of video games. Therefore, a walkthrough of the virtual environment is done using controls similar to typical video games, i.e., dragging with the mouse to look around and walking through the model with the arrow keys or the WASD keys on the keyboard on the user side.

### 5.3. Mission planning

A mission refers to the list of objectives for the robot. It consists of the locations where the robot will capture 360° pictures as well as coordinates of points on the path that the robot will follow to reach its goals. Once the user has selected various locations where the images should be captured, the robot's position in the BIM model is estimated from the position of the closest fiducial with respect to the robot's cameras and the known location of the fiducial in the BIM. The client interface transforms the location of the robot from the fiducial coordinate system (origin at the center of the fiducial) to the world coordinate system in Unity and visualizes the digital twin of the robot in the 3D virtual environment as shown in Figure 12.

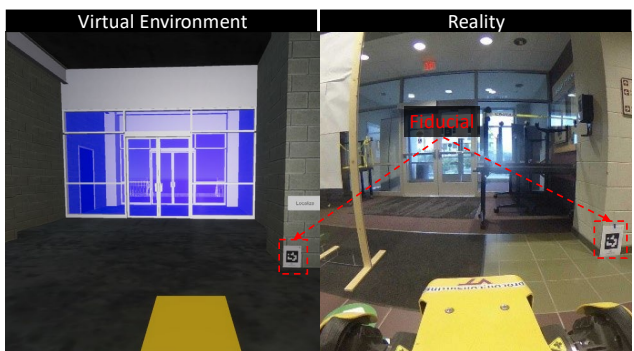

Figure 12 Fiducials in the virtual environment and in the real world

After localization, the system calculates the optimum path of the robot through all the DRPs using the greedy approach, i.e., iteratively finding the next closest DRP until all DRPs are reached. Here, the path is optimized for distance while avoiding obstacles like furniture and walls by deciding the order of DRPs to visit and finding the shortest walkable distance between the consecutive DRPs. After calculating the path, the cartesian coordinates of all the corners on the path are serialized into a string in the JSON format. The serialized information of the calculated path is transferred from the client interface to the middleware through a WebSocket connection relayed through the Cloud Server. The points on the path where the DRPs are located are also marked in the path information.

### 5.4. Robot control and Reality capture

After receiving the calculated path, the middleware device queries the robot's to estimate its position with respect to the closest visible fiducial. Then, the position of the next point on the path is transformed from the coordinate system centered at the fiducial to a coordinate system centered at the robot. This provides a vector from the robot's current location to the next point on the path. Then, the API of the robot is called to move the robot to the next point. For reality capture, a Ricoh Theta V 360° camera was used in this study. The 360° camera was installed on top of the robot using buckle clips with an adhesive base. When the robot reaches a DRP, the camera is triggered to capture a 360° image. The captured image is downloaded from the camera and stored in memory until the mission is completed to send back to the client interface for visualization. Once the mission is complete, all the downloaded images are sent back from the middleware to the client interface over another WebSocket connection.

### 5.5. Visualization

The client interface then iterates over all the captured 360° images and applies them as textures to the spheres representing the DRPs in the order of their presence on the last calculated path. Finally, the user navigates the 3D virtual environment to view the captured reality images as well as the pre-loaded BIM for progress monitoring. The application developed in Unity displays the interface shown in Figure 13, where each sphere in the 3D virtual environment is a 360° spherical image captured by the robot from the corresponding location within the job site. These 360° images can be accessed in real-time by inspectors and project stakeholders on the platform after the robot has completed its autonomous mission. This enables stakeholders to virtually navigate the site and view specific areas in detail, enhancing their ability to make informed decisions and manage the project more effectively.

The CPS also combines the schedule and floor plan of the building. Users can switch between the model view, floor plan view, and schedule view by clicking on the Floor plans (top right) and Schedule (bottom left) buttons in the interface shown in Figure 13. This allows the project stakeholders to view these different documents in one interface. The interface also allows switching inspection dates using the top left dropdown (Figure 13). Using this feature, project stakeholders can go back in time and see when an error occurred first time.

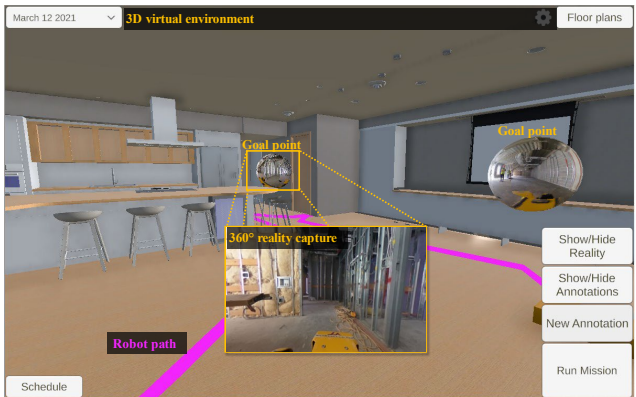

Figure 13 Visualization of 360° reality capture in the 3D virtual environment in Unity

## 6. Results

### 6.1. Experimental Investigation

This section presents the results of an experiment that was conducted to evaluate the prototype implementation of the proposed framework for correctness. The developed application was tested in Bishop-Favrao Hall (BFH) on Virginia Tech's campus in Blacksburg, VA, USA. The test mission involved walking through a space of approximately 4,841 sq.ft. in size, which included two classrooms and the entrance lobby, and capturing 360° images at six DRPs, as illustrated in Figure 8. The total length of the robot's path was 139.1 ft (42.4 m), and the robot successfully executed the mission within 3 minutes and 50 seconds with the moving speed set to 0.4m/s. The robot captured and sent back six 360° images for visualization in the 3D virtual environment.

The experiment revealed that installing fiducials at the site true to their corresponding position in the 3D model was challenging. Small residual errors in the orientation of the fiducials led to large deviations in the navigation of the robot as it moved farther from the fiducials. To address this challenge, more fiducials were installed to keep the deviations small. The required number of fiducials depends on the length of the path, and in this experiment, installing a fiducial every 8m on the robot's path ensured that at least one fiducial was always visible to the robot. Therefore, five fiducials were installed for the whole path at an average interval of 8 meters.

During the investigation, the robot's navigation through the door openings was found to be challenging due to the narrow clearance. The path occasionally overlapped with the walls due to the errors in the placement of the fiducials as well as the localization error arising from the motion of the robot. To improve the robot's navigation through the doorways, a fiducial was placed near the doors.

Figure 6 shows the 2D floor plan of BFH with the DRPs and the path of the test mission in the experiment. The results of this experiment demonstrate the consistency of the prototype with the proposed framework for robot navigation in indoor environments of construction sites. However, further improvements are necessary to overcome the challenges identified during the experiment.

### 6.2. Expert Validation

The expert validation was conducted with seventeen construction experts. Similar sample sizes were used for semi-structured expert interviews by other construction researchers [97,102–104]. Also, as can be seen from Figure 15, the cumulative average values of the factors Perceived Usefulness (PU) and Perceived Ease-Of-Use (PEOU) measured using the questionnaire at the end of the interviews also saturated, i.e., changed no more than 3% with each participant for last 10 participants. Therefore, further recruitment of experts was stopped due to saturation, which means adding more experts to the participant pool would not influence the results significantly [104]. The demographics of the included experts are shown in Figure 14.

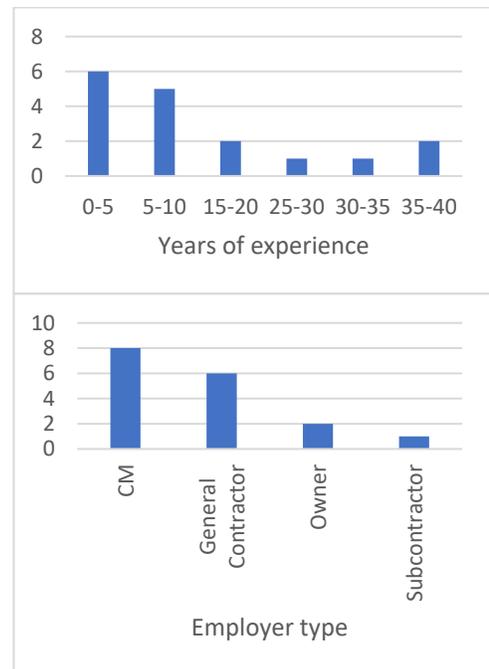

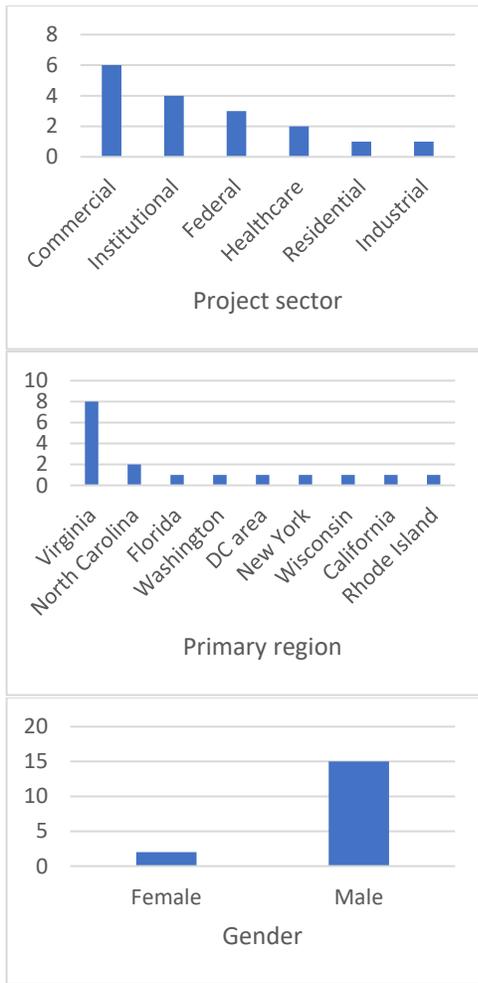

Figure 14 Demographics of interview participants

The average score for each of the questionnaire items is shown in Table 2. The questionnaire items were measured on a scale of 1 to 7. The questionnaire is provided in Appendix A. The results show high perceived usefulness and perceived ease of use among the research participants towards the robotic CPS developed in this study. The p-value of the hypothesis that the average score is more than 4 (mid-point of the scale) is less than 0.05. Hence, the null hypothesis that the average score is not more than 4 can be rejected with a confidence level of more than 95%. The perceived usefulness and perceived ease of use are the factors of acceptance according to the TAM. Therefore, it can be argued that there is a high acceptance of the proposed robotic CPS among the construction professionals interviewed during the research. Cronbach's alpha (α) is the indicator of the internal reliability of the scale used. An α-value of more than 0.7 is considered acceptable.

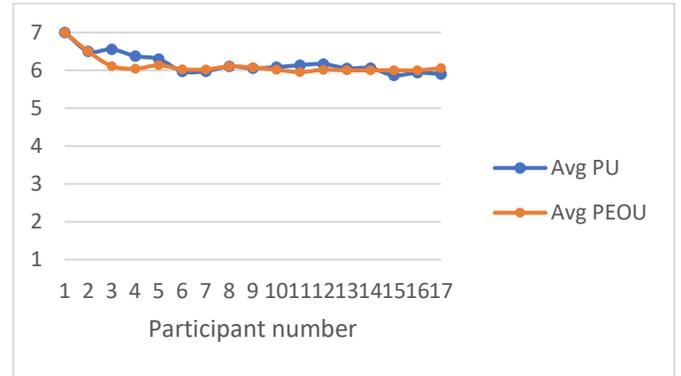

Figure 15 Cumulative PU and PEOU value

### 6.3. Factors affecting Perceived Usefulness

The first six questionnaire items measured the perceived usefulness of the system to the experts. It measures how much the experts believe that the given system will improve their ability to perform a task using the system [93]. The average score of the items pertaining to the perceived usefulness is 5.94 on a scale of 1 to 7, where 1 means strongly disagree and 7 means strongly agree. The interview results identified the factors that affected the perceived usefulness of the system either positively or negatively. In Figure 16, the results of the semi-structured interviews are presented. In the figure, the cells represent the codes generated from the analysis of the interview transcripts. These codes are clustered according to its impact on the perceived usefulness identified through content analysis. The vertical axis in the figure classifies the factors as having positive impact or negative impact. The horizontal axis represents whether these factors are present in the proposed CPS framework.

Table 2 Validation questionnaire results

| Factor | Questionnaire Item | Average score (out of 7) | Combined score | p-value (μ>4) | Cronbach's α | |
|---|---|---|---|---|---|---|
| Perceived Usefulness (PU) | Using the CPS in my job would enable me to accomplish tasks more quickly. | 5.94 | | | | |
| | Using the CPS would improve my job performance. | 5.71 | | | | |
| | Using the CPS in my job would increase my productivity. | 5.88 | 5.90 | <0.001 | 0.94 | |
| | Using the CPS would enhance my effectiveness on the job. | 5.88 | | | | 0.88 |
| | Using the CPS would make it easier to do my job. | 5.65 | | | | |
| | I would find the CPS useful in my job. | 6.35 | | | | |
| | Learning to operate the CPS would be easy for me. | 6.18 | | <0.001 | 0.90 | |
| | I would find it easy to get the CPS to do what I want it to do. | 5.82 | | | | |

| | | | |
|---|---|---|---|
| Perceived Ease-of-Use (PEOU) | My interaction with the CPS would be clear and understandable. | 5.88 | 6.06 |
| | It would be easy for me to become skillful at using the CPS. | 6.18 | |
| | I find the CPS clear and understandable | 6.12 | |
| | I would find the CPS easy to use. | 6.18 | |

### 6.3.1. Positive factors

Experts believe that each of the factors mentioned can positively impact the perceived usefulness of the robotic CPS (Cyber-Physical System) in the following ways:

1. **Up-to-date model:** An up-to-date model enhances the perceived usefulness of the robotic CPS by providing inspectors with the most accurate and current representation of the subject being inspected. This allows them to make informed decisions based on the latest information and ensures that any changes or issues are identified promptly.
2. **Safer for inspectors:** Enhancing the safety of inspectors through the use of robotic CPS is seen as a valuable aspect. By minimizing or eliminating the need for inspectors to physically be present in hazardous environments, the perceived usefulness of the system increases as it reduces potential risks to human life and well-being.
3. **Remote inspection saves time:** The ability to conduct inspections remotely saves time by eliminating the need for travel and on-site visits. This increased efficiency in the inspection process allows inspectors to allocate their time more effectively, leading to higher productivity and a perceived increase in the usefulness of the robotic CPS.
4. **Remote collaboration:** Experts noted that the robotic CPS can allow multiple people to look at the project together remotely which can be useful for collaboration.
5. **Managing multiple sites:** Remote access to project information and updates through the robotic CPS proves highly valuable, especially in the context of COVID and managing projects across different locations. It enables project managers to efficiently monitor multiple projects and gain visibility into site progress without the need for frequent physical visits.
6. **Frequent monitoring:** Frequent monitoring facilitated by the robotic CPS allows for regular data collection, analysis, and evaluation. This ensures that any changes or anomalies are detected promptly, leading to timely interventions and preventive actions. The perceived usefulness of the system is increased as it enables proactive monitoring and maintenance, reducing the likelihood of critical failures or incidents.
7. **Reduces the need for multiple people to go to the site:** By reducing or eliminating the requirement for multiple inspectors to physically visit the site, the robotic CPS minimizes logistical complexities and costs. This streamlining of resources and personnel enhances the perceived usefulness of the system by making the inspection process more efficient and cost-effective.
8. **Planned and as-built together:** Combining planned and as-built data in a unified manner enables a better understanding of the project.
9. **Different documents in one place:** Having floor plans, schedules, and 3D models integrated into a single platform in the robotic CPS allows for a better understanding of project progress and convenient access to relevant information, saving time and improving efficiency.
10. **Consistency of data capture**: The proposed robotic CPS can collect data more consistently from the same location.
11. **Combined models:** The integration of different models like architectural, structural, etc. is useful for providing a holistic view of the project.
12. **Better sense of place:** Experts noted that the 3D virtual environment provides a better sense of the project as it visualizes the project in 3D than through a combination of multiple 2D drawings.
13. **Ability to go back in time:** The ability to access and review past inspection data allows inspectors to conduct retrospective assessments, i.e., identify when an error started in the past, resulting in improved insights and corrective actions.
14. **More data:** Experts highlighted the need for more data can be provided through the CPS to make it more useful. Some examples of additional data that can be provided are detailed drawings, including sections and elevations. Experts noted that adding sensors to the CPS to monitor moisture, temperature, and toxic gases can also be useful. The ability to take measurements from the interface remotely was also mentioned as a needed feature by multiple experts.
15. **Integration of schedule with the BIM model**: Although the CPS provides the 3D model, schedule, and floor plan, they are not integrated with each other. Experts noted that it will be more useful if the schedule and model were integrated, i.e., selecting an object in the 4D BIM would show its scheduled dates, or selecting an activity in the schedule would highlight the corresponding elements on the BIM.

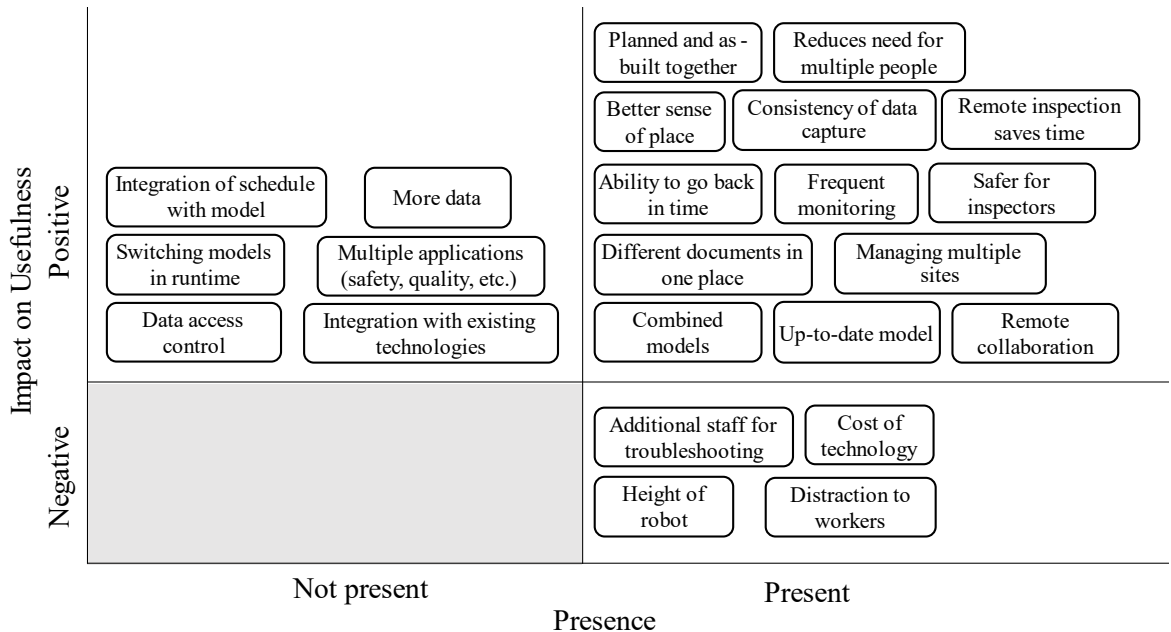

Figure 16 Factors affecting the perceived usefulness of the robotic CPS identified from the semi-structured interviews

16. **Switching models in runtime:** Experts noted that the level of details (LOD) in the BIM model required during different phases of the project can differ. For example, the BIM model used during the structural construction can differ from the one used during interior finishing. The CPS should allow changing the model in runtime.
17. **Multiple applications (safety, quality, etc.):** The ability of the robotic CPS to serve multiple applications, such as safety and quality inspections, enhances its usefulness. By accommodating diverse inspection needs, the system provides inspectors with a versatile toolset to address different requirements effectively. This flexibility increases the value and utility of the robotic CPS.
18. **Integration with existing technologies:** Seamless integration with existing technologies like Revit and Procore commonly used in construction will enhance the usefulness of the robotic CPS. By leveraging established infrastructures and workflows, inspectors can easily incorporate the system into their existing processes. This integration minimizes disruptions, improves efficiency, and maximizes the benefits derived from the robotic CPS.
19. **Data access control:** Robust data access control mechanisms increase the perceived usefulness of the robotic CPS by ensuring data security, privacy, and integrity. Inspectors can have confidence in the system's ability to protect sensitive information and control access based on user roles and permissions. This enhances trust in the system and encourages its adoption for critical inspection tasks.

The above factors contribute to the perceived usefulness of the robotic CPS by providing inspectors with enhanced capabilities, insights, and efficiency in their inspection processes.

### 6.3.2. Negative factors

The following factors were highlighted by the research participants that negatively affect the usefulness of the robotic CPS that must be addressed for increased adoption:

1. **Height of robot:** Construction experts emphasized the need for flexibility in capturing different elevations and perspectives, such as raising the camera or using telescopic options. As the robot used in this study has a maximum 33 inches of height, it may not capture enough details from the ceiling elements. The robot may also not capture details from the locations lower than the camera height, e.g., under kitchen cabinets.
2. **Distraction to workers:** One of the experts highlighted the possibility of distraction to the workers on site caused due to the operation of the robot. However, they also noted that the distraction may reduce over time as the workers get used to the robot.
3. **Cost of technology:** The cost of the technology was identified as a factor negatively impacting the usefulness. The associated cost of training people to operate the technology adds to the cost of the technology.
4. **Additional staff for troubleshooting:** One of the experts noted that to implement the CPS on the site, additional staff may be required to troubleshoot the robot which may reduce the usefulness of the CPS.

## 6.4. Factors affecting Perceived Ease-of-use

The last six questionnaire items in the TAM questionnaire measure the perceived ease of use of the system by the experts. It measures how much the experts find it easy to use the system to perform a task using the system [93]. The average score of the items pertaining to the perceived ease-of-use factor is 6.0 on a scale of 1 to 7. Figure 17 shows the results of the interview conducted with experts and lists the factors that affected the perceived ease-of-use of the system positively or negatively.

### 6.4.1. Positive factors

Experts believed that following factors would positively affect the ease-of-use of the proposed robotic CPS:

a) **Simple controls:** Experts noted that the simple control of the system using few mouse clicks and key presses made the system easy-to-use. The interface was also designed using the metaphor of video games. People who were familiar with video games noted that it was very easy to understand the controls.
b) **Flexible controls:** Some experts also noted that it is important to have flexibility in the control of the system in terms of speed of movement, inverse control of mouse movement. Also, the flexibility to select the point of inspection was mentioned to be an important feature as the regions of interest change from day to day as the construction progresses.
c) **Easy to share:** One of the experts noted that the ability to share the project data with others using simple URL was very important will make it easy to collaborate with others.
d) **Ease of learning:** Many experts noted that the system is easy to learn and they could quickly use it for performing progress monitoring. Ease of learning was highlighted as a key factor of ease of use as people don't want to spend a lot of time in learning a new system even if it saves them time in the long run.
e) **Bird-eye view:** The bird eye view of the project through the floor plan was found to be an important feature as it allowed the experts understand the whole space and what they were looking at currently.
f) **Subcontractor cooperation:** Cooperation of the trades subcontractors was highlighted by multiple experts. They noted that many experienced tradespeople would not like the presence of new technologies. One of the experts noted that often different tradespeople do not like to cooperate with each other and may cause hindrance in the operation of the robot.
g) **Similar controls as other software:** Few experts mentioned that they were confused with the controls because they used to a different control in other software like Revit or Navisworks. They highlighted that using similar controls that people are already used to may improve the ease of learning of the system.
h) **Seeing documents and reality at same time:** Although the interface of the robotic CPS allows switching between schedule, model, floor plan, and reality capture, some experts noted that they would like to view some of the documents at the same time, e.g., floor plan and model while navigating, schedule and reality capture for comparison of progress.
i) **Easier transition between model and reality:** A few experts found themselves disoriented while switching between the model and the reality capture and sometimes switched between views inadvertently. They highlighted the need for smoother transition

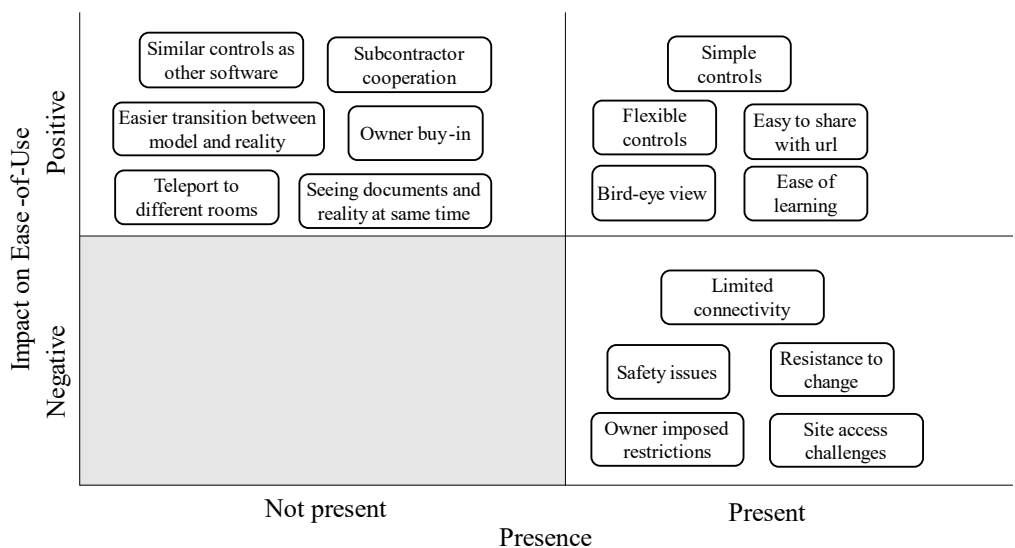

Figure 17 Factors affecting the perceived ease of use of the robotic CPS identified from the semi-structured interviews

between the different views and providing feedback from the system before and after the transition.

j) **Owner buy-in:** Experts noted that owners also need to understand and see the value of the technology before implementing it. Safety and legal risks are important considerations, and getting project owners to accept and address these risks upfront is crucial for successful adoption.

k) **Teleport to different rooms:** Some experts desired easier navigation through the model. They preferred navigating through the project structure by looking at it from a bird eye view on a floor plan and clicking on the floor to directly go to a room instead of manually navigating through the building.

### 6.4.2. Negative factors

The following negative factors were identified by the interviewed experts that reduce their perceived ease of use:

a) **Site access challenges:** The accessibility of the construction site for the robot is a major issue that limits the ease of use of the robotic CPS. Many experts highlighted that the construction site is a dynamic environment that changes continuously with new components being installed and new materials being delivered every day. Moving the robot from one floor to another when there are no permanent stairs is also a major barrier to the use of the CPS. Also, the CPS relies on the use of fiducials for robot navigation that must be installed on the site. Maintaining those fiducials is also challenging. Providing charging points for the robot was also identified as an important concern.

b) **Limited connectivity:** Some experts noted that Wi-Fi and internet connectivity are not available at the site during the initial period, which is required for the operation of the CPS.

c) **Safety issues:** There are many safety issues with the use of robots at the site. For example, one of the experts noted that it may cause accidents if people do not see it coming.

d) **Resistance to change:** The implementation of the robotic CPS may be challenging due to resistance to learning new things by the construction workforce. As one of the experts noted, learning to work with the new technology may be seen as an additional workload by experienced construction personnel. People may want to do the work in the traditional ways if they find doing it that way easier than learning a new workflow.

e) **Owner-imposed restrictions:** Some experts who have worked on government projects mentioned that the owner being the federal or state governments sometimes impose restrictions on the use of cameras or other types of devices due to security risks on certain parts of the project. This may make it difficult for the robotic CPS to be used.

## 7. Discussion

The acceptance study conducted with construction professionals shows that there is a high level of acceptance of the system by the professionals. Although the experts interviewed during the study showed acceptance of the system, they also revealed many critical success factors of the system. The most important success factor identified by the experts is related to the cost. Owner and subcontractor buy-in are also important factors identified from the qualitative analysis of the interviews. Site accessibility which includes the ability of the robot to access different parts of the site as well as the ability to connect to the internet onsite during construction was also highlighted by many experts. Therefore, these factors must also be addressed while deploying the proposed robotic CPS for progress monitoring.

The interview results identified that the remote aspect of the CPS that allows capturing site information without traveling to the site and providing the information to multiple stakeholders is an important contributor to the usefulness of the CPS for progress monitoring. The availability of different project documents like BIM, schedule, and floor plans in a single interface is also useful for progress monitoring.

One of the major challenges in implementing CPS in construction is the integration of the digital models/components with the physical construction [63], i.e., creating a robust cyber-physical bridge. Other technological challenges of adopting CPS in construction include the development of an interoperable and reliable system and also managing data privacy [89]. Cyber-Physical Systems are subject to cyber-security threats that put the data privacy of construction projects at risk [73]. There are other educational challenges related to re-training the workforce and legal challenges related to the security and liability of the CPS [89].

## 8. Conclusion

This article presents a conceptual framework for a robotic cyber-physical system for automated reality capture and visualization in construction progress monitoring. The integration of a quadruped robot, Building Information Modelling, and 360° camera provides a high-resolution and real-time visualization of the construction progress, which can be used for progress monitoring and communication between project stakeholders. The experimental investigations were conducted to evaluate a prototype of the proposed CPS framework, and the acceptance of the CPS was evaluated through semi-structured interviews with industry experts involved in progress monitoring work. The findings indicate that the proposed CPS framework is useful for the progress monitoring process by capturing as-built data from the site autonomously and providing different information like the BIM model, floor plan, schedule, and reality capture together in one interface.

This study makes several contributions to the body of knowledge as well as practice. First, it developed a robotic CPS framework for automated reality capture and visualization. Second, it generated knowledge of factors that affect perceived usefulness and perceived ease-of-use of the robotic CPS that ultimately affect the acceptance of the technology.

The proposed system also has some limitations. Firstly, the current system is proposing a fundamental framework for automated reality capture and progress monitoring using a robotic CPS. More detailed implementation and on-site investigation is in the scope for future work. The proposed framework with quadruped robot may not yet be suitable for all types of construction sites, such as those with complex terrains or areas that are difficult for the robot to access due to the current limitations of robotic technologies. Secondly, the system's accuracy may be affected by environmental factors such as lighting, weather conditions, and obstacles on the construction site. Finally, the initial cost of the system may be a barrier for some construction management teams.

There is a need for further investigation into improving the accuracy of the system under various environmental conditions and the development of a more affordable system that can be adopted by construction management teams of various sizes. Additionally, research can explore the use of machine learning and artificial intelligence to improve the system's capabilities, such as the ability to identify and track specific objects or materials on the construction site and re-adjust its path due to new obstacles introduced at the site. Currently, the robot's path is optimized for distance only. However, a multi-objective optimization may be performed to avoid areas with higher human traffic and minimize distractions to workers in busy areas.

Overall, the proposed robotic CPS system and the findings from the interviews have the potential to bring significant benefits to the construction progress monitoring process. Construction management teams may consider adopting this technology to improve progress monitoring and communication with stakeholders, ultimately leading to more efficient and successful construction projects.

## Acknowledgements

This study was partially funded by Industrialized Construction Initiative at Virginia Tech College of Engineering. Additionally, the authors acknowledge the work of undergraduate student Kalen Rita for supporting this research.